# Gyroscopically Stabilized Robot: Balance and Tracking

**Yongsheng Ou & Yangsheng Xu**
The Chinese University of Hong Kong
Department of Automation and Computer-Aided Engineering
you,ysxu@acae.cuhk.edu.hk

*Abstract:* The single wheel, gyroscopically stabilized robot - Gyrover, is a dynamically stable but statically unstable, underactuated system. In this paper, based on the dynamic model of the robot, we investigate two classes of nonholonomic constraints associated with the system. Then, based on the backstepping technology, we propose a control law for balance control of Gyrover. Next, through transferring the systems states from Cartesian coordinate to polar coordinate, control laws for point-to-point control and line tracking in Cartesian space are provided.
**Keywords**: Nonholonomic constraints, robot control, nonlinear control, underactuated systems, dynamically stabilized robot, mobile robot

## 1. Introduction

Recently, there has been growing interest in the design of feedback control laws for nonholonomic systems (Kolmanovsky, I. & McClamroch, N. H. (1995)). Due to Brockett's theorem in (Zabczyk, J. (1989)), it is well known that a nonholonomic control system cannot be asymptotically stabilized to a resting configuration by smooth time-invariant control laws (Bloch, A. M.; Reyhanoglu, M. & McClamroch, N. H. (1992)). Despite this, several discontinuous or time-variant approaches have been proposed for stabilizing such systems in (Bloch, A. M.; Reyhanoglu, M. & McClamroch, N. H. (1992)), (Indiveri, G. (1999)), (Tayebi, A. & Rachid, A. (1997)), (Lee, T. C.; Song, K.T.; Lee, C.H. & Teng, C. C. (2001)) and (Kolmanovsky, I. & McClamroch, N. H. (1995)). The references above refer to systems with first-order nonholonomic constraints, which can usually be expressed in terms of nonintegrable linear velocity relationships. However, there is another kind of mobile robot systems, which possess both first-order and second-order nonholonomic constraints, such as our robot -- Gyrover, bicycles and motorcycles. Mobile robots have their inherent nonholonomic features, which can be described as first-order nonholonomic constraints among joint velocities and Cartesian space velocities. These arise when robots roll on the ground without slipping. Because no actuators can be used directly for stabilization in the lateral direction, these systems are underactuated nonlinear systems. This induces another nonholonomic constraint of robots. Thus, to compare with these above research works, our mobile robot systems -- Gyrover, is more challenging to be controlled.

This paper can provide some ideas for that class of problems. In this paper, we want to control an underactuated mobile robot system -- Gyrover. There are two control inputs: one is the steering torque (or rate) and the other is the driving torque (or speed). However, we have four independent generalized coordinates to control: (1) the lean angle, (2) the heading angle, (3) the Cartesian space $X$ axis, (4) the Cartesian space $Y$ axis.
Our previous papers (Au, K. W. & Xu, Y. (1999)) and (Au, K. W. & Y. Xu, (2000)) assume that the robot remains around the vertical position, which simplify this nonlinear system to a linear one. Therefore validation of the results can be limited.
Some work has been on the tracking of a rolling disk, such as in (Rui, C. & McClamroch, N. H. (1995)), which assumed three control inputs in the direction of steering, leaning and rolling, where no unactuated joint and no second-order constraint is presented. In (Getz, N. H. (1995)), the author simplified the bicycle dynamic model, and used velocities as control inputs to enable the lean angle (called "roll-angle" in that paper) to track trajectories, which have continuous differentials. However, the controller could not guarantee that the bicycle would not topple over, i.e. the lean angle was out of range, before convergence.
In this paper, we focus on three control problems that have not yet been solved for this robot. The first problem is the balance of the robot while standing. The second problem is concerned with point to point control. The third problem relates to following a straight line. These three problems are of significance in controlling a system with both first-order and second-order nonholonomic constraints.

## 2. Gyrover

We have developed a single wheel, gyroscopically stabilized robot, Gyrover, over several generations, each one more sophisticated, reliable and capable of better performance (Brown, H. B. & Xu, Y. (1996)). Figure 1 shows a photograph of Gyrover.

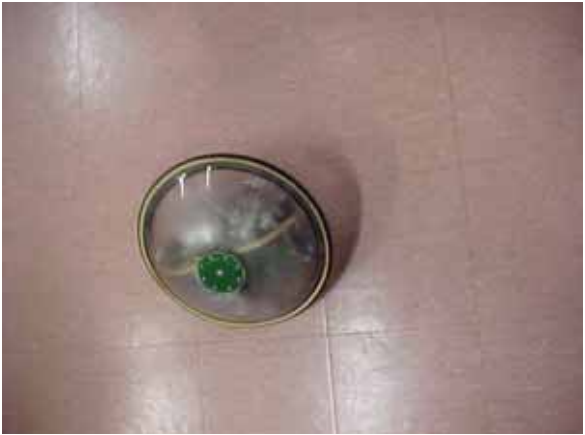

Fig. 1. A single wheel robot, Gyrover

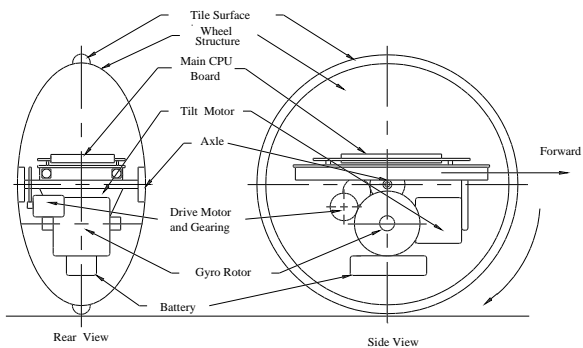

Fig. 2. The basic configuration of Gyrover

Gyrover is a novel, single wheel gyroscopically stabilized robot. Figure 2 shows a schematic representation of the mechanism design. In brief, the robot is a sharp-edged wheel, inside of which an actuation mechanism is fitted. The actuation mechanism consists of three separate actuators: (1) a spin motor, which spins a suspended flywheel at a high rate, imparting dynamic stability to the robot; (2) a tilt motor, which controls the steering of the robot; and (3) a drive motor, which causes forward and/or backward acceleration, by driving the single wheel directly.

The behavior of Gyrover is based on the principle of gyroscopic precession as exhibited in the stability of a rolling wheel. Because of its angular momentum, a spinning wheel tends to precess at right angles to an applied torque, according to the fundamental equation of gyroscopic precession:

$$T = J\omega \times \Omega$$

where $\omega$ is the angular speed of the wheel, $\Omega$ is the wheel's precession rate, normal to the spin axis, $J$ is the wheel polar moment of inertia about the spin axis, and $T$ is the applied torque, normal to the spin and precession axes. Therefore, when a rolling wheel leans to one side, rather than falling over, the gravitationally induced torque causes the wheel to precess so that it turns in the direction that it is leaning. The robot supplements this basic concept with the addition of an internal gyroscope -- the spinning flywheel -- nominally aligned with the wheel and spinning in the direction of forward motion. The flywheel's angular momentum produces lateral stability when the wheel is stopped or moving slowly.

Gyrover has a number of potential advantages over multi-wheeled vehicles:

1. The entire system can be enclosed within the wheel to provide mechanical and environmental protection for equipment and mechanisms.
2. Gyrover is resistant to getting stuck on obstacles because it has no body to hang up, no exposed appendages, and the entire exposed surface is *live* (driven).
3. The tiltable flywheel can be used to right the vehicle from its statically stable, rest position (on its side). The wheel has no "backside" on which to get stuck.
4. Gyrover can turn in place by simply leaning and precessing in the desired direction, with no special steering mechanism, enhancing maneuverability.
5. Single-point contact with the ground eliminates the need to accommodate uneven surfaces and simplifies control.
6. Full drive traction is available because all the weight is on the single drive wheel.
7. A large pneumatic tire may have very low ground-contact pressure, resulting in minimal disturbance to the surface and minimum rolling resistance. The tire may be suitable for traveling on soft soils, sand, snow or ice; riding over brush or other vegetation; or, with adequate buoyancy, for traveling on water.

Potential applications for Gyrover are numerous. Because it can travel on both land and water, it may find amphibious use on beaches or swampy areas, for general transportation, exploration, rescue or recreation. Similarly, with appropriate tread, it should
travel well over soft snow with good traction and minimal rolling resistance. As a surveillance robot, Gyrover's slim profile enables it to pass through doorways and narrow passages, and its ability to turn in place to maneuver in tight quarters. Another potential application is as a high-speed lunar vehicle, where the absence of aerodynamic disturbances and low gravity would permit efficient, high-speed mobility. As Gyrover's development progresses, we anticipate that other more specific uses will become evident.

## 3. Inertia Matrix and Nonholonomic Constraints

The kinematics and dynamics of Gyrover are different from those of unicycles, such as in (Aicardi, M.; Casalino, G.; Balestrino, A. & Bicchi, A. (1994)). The difference lies in the assumption that the unicycle always remains vertical. On the contrary, Gyrover can be considered as a rolling disk which is not constrained to the vertical position and is connected to a high speed spinning flywheel. This model can serve well as a simplification for the study of a model of Gyrover.

Consider a disk rolling without slipping on a horizontal plane as shown in Figure 3. Let $\sum_o X,Y,Z$ and $\sum_c x,y,z$ be the inertial frame whose $x-y$ plane is anchored to the flat surface and the body coordinate frame whose origin is located at a center of the rolling disk, respectively. Let $(X,Y,Z)$ be the Cartesian coordinates of the center of mass $(c)$ with respect to the inertial frame $\sum_o$. Let $A$ denotes the point of the contact on the disk. The configuration of the disk can be described by six generalized coordinates $(X,Y,Z,\alpha,\beta,\gamma)$, where $\alpha$ is the steering (precession) angle measured from $X$-axis to the contact line, $\beta \in (0,\pi)$ is the lean (nutation) angle measured from $Z$-axis to the $z$, and $\gamma$ is the rolling angle. $R$ is the radius of Gyrover. $m, I_x, I_y, I_z$ represent the total mass and the moment of inertia of Gyrover.

In the derivation of the model, we assume that the wheel rolls on the ground without slipping. Based on the previous derivation in (Xu, Y.; Brown, H.B. & Au, K. W. (1999)) and letting $S_x := \sin(x)$, $C_x := \cos(x)$, the dynamic model is as follows.

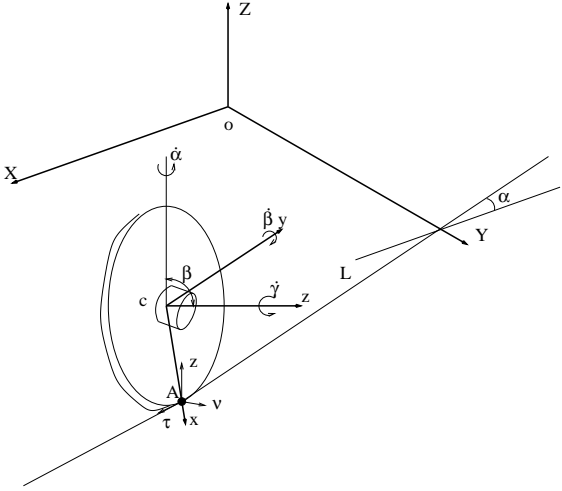

Fig. 3. System parameters of Gyrover's simplified model

$$M(q)\ddot{q} = N(q,\dot{q}) + Bu \quad (1)$$

$$\begin{cases} \dot{X} = R(\dot{\gamma}C_\alpha + \dot{\alpha}C_\alpha C_\beta - \dot{\beta}S_\alpha S_\beta) \\ \dot{Y} = R(\dot{\gamma}S_\alpha + \dot{\alpha}S_\alpha C_\beta + \dot{\beta}C_\alpha S_\beta) \end{cases} \quad (2)$$

where $q = [\alpha, \beta, \gamma]^T$, $N = [N_1, N_2, N_3]^T$

$$M = \begin{bmatrix} M_{11} & 0 & M_{13} \\ 0 & M_{22} & 0 \\ M_{31} & 0 & M_{33} \end{bmatrix}$$

$$B = \begin{bmatrix} 1 & 0 & 0 \\ 0 & 0 & 1 \end{bmatrix}^T, u = \begin{bmatrix} u_1 \\ u_2 \end{bmatrix}$$

$M_{11} = I_x S_\beta^2 + (2I_x + mR^2)C_\beta^2$

$M_{13} = (2I_x + mR^2)C_\beta$

$M_{31} = M_{13}$

$M_{33} = 2I_x + mR^2$

$N_1 = (I_x + mR^2)S_{2\beta}\dot{\alpha}\dot{\beta} + 2I_x S_\beta \dot{\beta}\dot{\gamma}$

$N_2 = -mgRC_\beta - (2I_x + mR^2)S_\beta \dot{\alpha}\dot{\gamma}$

$\quad - (I_x + mR^2)C_\beta S_\beta \dot{\alpha}^2$

$N_3 = 2(I_x + mR^2)S_\beta \dot{\alpha}\dot{\beta}$

where $(X,Y,Z)$ is the robot's center of mass coordinate with respect to the inertial frame as shown in Figure 3. $M(q) \in R^{3\times 3}$ and $N(q,\dot{q}) \in R^{3\times 1}$ are the inertial matrix and nonlinear terms, respectively. Equations (1) and (2) form the dynamic model and first-order nonholonomic constraints in the form of velocity.

In order to simplify the control design, we will transform the inertia matrix of the model to a diagonal matrix and reduce the nonlinear terms of the dynamic model. We first intend to cancel some nonlinear terms by letting

$$\begin{cases} u_1 = -N_1 + u_3 \\ u_2 = -N_3 + u_4. \end{cases} \quad (3)$$

Substituting Equation (3) into Equation (1) yields

$$\begin{cases} M_{11}\ddot{\alpha} + M_{13}\ddot{\gamma} = u_3 \\ M_{22}\ddot{\beta} = -mgRC_\beta - (I_x + mR^2)C_\beta S_\beta \dot{\alpha} \\ \quad\quad\quad - (2I_x + mR^2)S_\beta \dot{\alpha}\dot{\gamma} \\ M_{31}\ddot{\alpha} + M_{33}\ddot{\gamma} = u_4 \end{cases} \quad (4)$$

From the first and third equations in Equation (4) solving with respect to $\ddot{\alpha}$ and $\ddot{\gamma}$ one obtains

$$M_\rho = M_{11}M_{33} - M_{13}^2$$

Due to its "shell" structure, even when Gyrover topples over, the lean angle is not zero, so that $S_\beta^2 > 0$ and $M_\rho > 0$ for all $t$. Then, we let

$$\begin{cases} u_5 = (M_{33}/M_\rho)u_3 - (M_{13}/M_\rho)u_4 \\ u_6 = -(M_{13}/M_\rho)u_3 + (M_{11}/M_\rho)u_4 \end{cases}$$

and obtain

$$\begin{cases} \ddot{\alpha} = u_5 \\ M_{22}\ddot{\beta} = -mgRC_\beta - (I_x + mR^2)C_\beta S_\beta \dot{\alpha} \\ \quad\quad\quad - (2I_x + mR^2)S_\beta \dot{\alpha}\dot{\gamma} \\ \ddot{\gamma} = u_6 \end{cases} \quad (5)$$

Equation (2) is nonintegrable, which is defined in (Reyhanoglu, M.; van der Schaft, A. J.; McClamroch, N. H. & Kolmanovsky, I. (1999)). Hence, it is the first-order nonholonomic constraint of the robot system. Moreover, we note that no control input is available for

25

actuating directly on lean angle $\beta$. This forms a constraint in form of accelerations. G. Oriolo in (Oriolo, G. & Nakamura, Y. (1991)) proposed some necessary conditions for the partial integrability of second-order nonholonomic constraints, one of the conditions is the gravitational term $G_u$ is constant. In Equation (5), the gravitational term varies along $\beta$, thus it is a nonintegrable, second-order nonholonomic constraint.

## 4. Balance Control

Our first control problem is to enable the robot to stand vertically, i.e., to stabilize the lean angle $\beta$ to $\pi/2$ and $\dot{\beta}$, $\ddot{\beta}$, $\dot{\alpha}$ and $\dot{\gamma}$ to zero. Based on the previous considerations, we are now able to specify more clearly the aforementioned closed loop steering problem in the following general terms.

Let the robot system defined in Equation (5) be initially moving in an undesired state and assume all essential state variables are directly measurable. Then find a suitable (if any) state dependent control law $[u_5, u_6]^T$, which guarantees the states $[\beta - \pi/2, \dot{\beta}, \ddot{\beta}, \dot{\alpha}, \dot{\gamma}]^T$ to be asymptotically driven to the null limiting point $[0,0,0,0,0]^T$, while avoiding any attainment of the conditions of $\beta = 0$ ($\beta = \pi$) in any finite time.

Firstly, we let

$$\begin{cases} G_m = mg\,R/M_{22} \\ I_m = (I_x + mR^2)/M_{22} \\ J_m = (2I_x + mR^2)/M_{22} \end{cases}$$

where $G_m$, $I_m$ and $J_m$ are positive.

Then, the dynamic equations become

$$\begin{cases} \ddot{\alpha} = u_5 \\ \ddot{\beta} = -G_m C_\beta - I_m C_\beta S_\beta \dot{\alpha} - J_m S_\beta \dot{\alpha}\dot{\gamma} \\ \ddot{\gamma} = u_6 \end{cases} \quad (6)$$

Letting $x^{(i)}$ be the $i^{th}$ time derivative of $x$, then from Equation (6), we have

$$\beta^{(3)} = h_1(t)\dot{\beta} + h_2(t)u_5 + h_3(t)u_6 \quad (7)$$

where

$$\begin{cases} h_1(t) = G_m S_\beta - I_m C_{2\beta}\dot{\alpha}^2 \\ h_2(t) = -I_m S_{2\beta}\dot{\alpha} - J_m S_\beta \dot{\gamma} \\ h_3(t) = -J_m S_\beta \dot{\alpha}^2. \end{cases}$$

Let $\beta(0) - \pi/2 = a$, $\dot{\beta}(0) = b$, $\ddot{\beta}(0) = c$, and a real number $\sigma$ be

$$\sigma = |3a + 2b + c|/2 + |a + 2b + c|/2 + |a+b|/\sqrt{2}.$$

**Proposition 1** Consider the system (6) with the feedback control laws $u_5$ and $u_6$,

$$u_5 = \begin{cases} -(\dot{\alpha} - \sqrt[4]{k_2 V}) & if\ \dot{\alpha}(0) \\ -(\dot{\alpha} + \sqrt[4]{k_2 V}) & otherwise \end{cases} \quad (8)$$

where $V$ is defined in Equation (10) and $k_2$ is a positive number, which can be designed, and

$$u_6 = -(3(\beta - \pi/2) + 5\dot{\beta} + 3\ddot{\beta} \\ + h_1(t)\dot{\beta} + h_2(t)u_5)/h_3(t) \quad (9)$$

where, $h_1(t)$, $h_2(t)$, $h_3(t)$ are defined as in Equation (7).

Any state $(\dot{\alpha}, \dot{\gamma}, \beta, \dot{\beta}, \ddot{\beta})$ starting from the domain $D$, which is defined as

$$D = \{(\dot{\alpha}(0), \dot{\gamma}(0), \beta(0), \dot{\beta}(0), \ddot{\beta}(0)) | \dot{\alpha}(0) \neq 0, \\ 0 < \beta(0) < \pi, \sigma < \pi/2, \dot{\alpha}, \dot{\gamma}, \beta, \dot{\beta}, \ddot{\beta} \in R^1\}$$

converges to the limiting point $[0,0,\pi/2,0,0]^T$.

To present the proof, we need a lemma.

**Lemma 1** Consider a mechanical system with a constraint among a number of states $x_1(t), x_2(t),\ldots x_n(t) \in R^1$. One of these state variables is uniquely determined by the other states; i.e., $x_n = f(x_1, x_2 \ldots x_{n-1})$. Let the limit of $x_n$ exist as $t \to \infty$ and all of the other states be asymptotically stabilizable to some real values. If all of the states are continuous and bounded, for all of $t$, then, $x_n$ is also asymptotically stabilized to its limit, which is decided by the other states.

*Proof:*

First, since $x_n(t)$ is continuous and bounded, for all $t$, it is stable.

Second, because all of the other states are asymptotically stabilized to some values, thus

$$\lim_{t \to \infty} x_i = e_i \quad exist \quad i = 1,2,\ldots,n-1.$$

Moreover, according to the property of limits, we have

$$\lim_{t \to \infty} x_n = \lim_{t \to \infty} f(x_1, x_2,\ldots x_{n-1}),$$

$$\lim_{t \to \infty} x_n = f\left(\lim_{t \to \infty} x_1, \lim_{t \to \infty} x_2, \ldots \lim_{t \to \infty} x_{n-1}\right).$$

Because $x_n$ is uniquely decided by the other states and has a limit as time goes to infinity, $x_n$ will converge to the limit decided by the other states.

Then, we address the proof for Proposition 1.

*Proof:*

First, to prove the subsystem $\beta, \dot{\beta}, \ddot{\beta}$ asymptotically stabilized, we consider the following positive definite Lyapunov function candidate $V$, defined on $D$,

$$V = (\beta - \pi/2)^2/2 + (\dot{\beta} + \beta - \pi/2)^2/2 \\ \ddot{\beta} + 2(\dot{\beta} + \beta - \pi/2)^2/2. \quad (10)$$

We need to solve two more problems before we can prove the subsystem is asymptotically stable. One is whether $\dot{\alpha}$ is not zero in any finite time; the other is whether the controllers guarantee $\beta$ is constrained in $(0, \pi)$. We will address the first problem later. We prove the second problem by replacing $u_6$ into Equation (6). We have

$$\beta^{(3)} = -(3(\beta - \pi/2) + 5\dot{\beta} + 3\ddot{\beta}). \quad (11)$$

By solving this linear differential equation, we obtain
$$\beta - \pi/2 = e^{-t}((3a+2b+c) + \sqrt{2}(a+b)\sin(\sqrt{2}t)$$
$$-(a+2b+c)\cos(\sqrt{2}t)/2.$$
From $\sigma < \pi/2$, we know that $0 < \beta < \pi$ for all $t$.
Then, from Equation (10) the time derivative of V is
$$\dot{V} = (\beta - \pi/2) + (\dot{\beta} + \beta - \pi/2)(\ddot{\beta} + \dot{\beta})$$
$$+ (\ddot{\beta} + 2(\dot{\beta} + \beta - \pi/2))(\beta^{(3)} + 2(\ddot{\beta} + \dot{\beta})).$$
By substituting Equation (11) into it, we have
$$\dot{V} = (\beta - \pi/2) + (\dot{\beta} + \beta - \pi/2)(\ddot{\beta} + \dot{\beta})$$
$$+ 3(\ddot{\beta} + 2(\dot{\beta} + \beta - \pi/2))(\ddot{\beta} + \dot{\beta} + \beta - \pi/2).$$
Then,
$$\dot{V} = -(\beta - \pi/2)^2 - (\dot{\beta} + \beta - \pi/2)^2 \qquad (12)$$
$$-(\ddot{\beta} + 2(\dot{\beta} + \beta - \pi/2)).$$
From Equations (10) and (12), we have
$$\dot{V} = -2V \qquad (13)$$
We can use the following Lyapunov function to prove $\dot{\alpha}$ converging to zero.
$$V_\alpha = \sqrt{k_2 V}/4 + \dot{\alpha}^2/2.$$
The time derivative of $V_\alpha$ is
$$\dot{V}_\alpha = \frac{\sqrt{k_2}\dot{V}}{8\sqrt{V}} + \dot{\alpha}u_5.$$
By substituting Equations (8) and (13) into it, we have
$$\dot{V}_\alpha = -(\sqrt[4]{k_2 V}/2 \mp \dot{\alpha})^2. \qquad (14)$$
Then we prove that the condition $\dot{\alpha} = 0$ cannot ever be approached in any finite time. Without losing generality, we assume $\dot{\alpha}(0) > 0$ and let $k_2$ be 1.
If we let $V(0) = V_0$, from Equation (13), and by solving the differential equation, we obtain
$$V = e^{-2t}V_0.$$
By putting it into Equation (8}), we have
$$\ddot{\alpha} = -(\dot{\alpha} - e^{-t/2}\sqrt[4]{V_0}).$$
By solving this differential equation, we obtain
$$\ddot{\alpha} = e^{-t}\dot{\alpha}(0) + 2(e^{-t/2} - e^{-t})\sqrt[4]{V_0}.$$
Thus, since $\dot{\alpha}(0) > 0$, $\dot{\alpha}$ cannot reach zero in any finite time.
Since $S_\beta$ and $\dot{\alpha}$ are not zero in any finite time, then $h_3(t)$ does not become zero for all $t$ and from Equation (15), $\dot{\gamma}$ is continuous. We will prove that the limit of $\dot{\gamma}$ exists and is zero, as time goes to infinity.
From Lemma 1, if we asymptotically stabilize $\dot{\beta}, \beta, \dot{\alpha}$ and guarantee that all of these states are continuous and bounded, $\dot{\gamma}$ will asymptotically converge to zero. Since $\dot{\alpha}$ will never be zero, there are two problems left. One is whether $\beta, \dot{\beta}, \ddot{\beta}$ can be stabilized and the other is whether we can guarantee $0 < \beta < \pi$ for all of $t$. From Equation (6}), we have

$$\dot{\gamma} = \frac{\ddot{\beta} + G_m C_\beta + I_m C_\beta S_\beta \dot{\alpha}^2}{-J_m S_\beta \dot{\alpha}}. \qquad (15)$$

As $V$ and $\dot{\alpha}$ reach very small values, $\sqrt{V}$ is the higher order small of $\sqrt[4]{V}$. For $|\beta - \pi/2| \le \sqrt{V}$, $|\beta - \pi/2|$ is the higher order small of $\dot{\alpha}$, and so as to $\dot{\beta}$ and $\ddot{\beta}$. Using a Taylor series expansion, $C_\beta$ is a higher order small of $\dot{\alpha}$. Thus, from Lemma 1 and the above equation, $\dot{\gamma}$ asymptotically converges to zero.

## 5. Position Control

Here, we propose a controller that drives the robot to the Cartesian space origin to study the point to point control problem. This is extremely important, for it serves as the basis for Cartesian space tracking.

Since the origin of the frame $XYZO\left(\sum o\right)$ in Figure 3 is fixed on the ground, for the purposes of tracking problems, it is more direct to use the point of contact A on the ground to describe the position of the robot, instead of the center of mass. Let $(x_\alpha, y_\alpha)$ be the coordinates of the contact point A on the ground that coincides with a point of contact P of the robot in Figure 3. $x_a$ and $y_a$ can be expressed as

$$\begin{bmatrix} x_a \\ y_a \end{bmatrix} = \begin{bmatrix} X - RS_\alpha C_\beta \\ Y - RC_\alpha C_\beta \end{bmatrix} \qquad (16)$$

Differentiating Equation (16) with respect to time, we obtain

$$\begin{cases} \dot{x}_a = R\dot{\gamma}C_\alpha \\ \dot{y}_\alpha = R\dot{\gamma}S_\alpha \end{cases} \qquad (17)$$

There are two kinds of input control commands for Gyrover: one set of control commands are torques and the other set of commands are velocities. The velocity commands $u_\alpha$ and $u_\gamma$ control $\dot{\alpha}$ and $\dot{\gamma}$, respectively. Thus, Equations (2) and (5) transform into

$$\begin{cases} \dot{\alpha} = u_\alpha \\ \ddot{\beta} = -G_m C_\beta - I_m C_\beta S_\beta \dot{\alpha} - J_m S_\beta \dot{\alpha}\dot{\gamma} \\ \dot{\gamma} = u_\gamma \\ \dot{x}_a = Ru_\gamma C_\alpha \\ \dot{y} = Ru_\gamma S_\alpha \end{cases} \qquad (18)$$

Let us consider the robot with respect to the inertial frame $XYZO\left(\sum o\right)$, as shown in Figure 4.

By representing the Cartesian position of the robot in terms of its polar coordinates, which involves the error distance $e \ge 0$, measured from A to O (the origin of the frame), Gyrover's orientation $\theta$ with respect to the inertial frame, and defining $\psi = \theta - \alpha$ as the angle measured between the vehicle principal axis and the distance vector $e$. When $e = 0$, there is no definition for $\theta$ and $\psi$. Then the following equations are obtained

$$\begin{cases} \dot{e} = Ru_\gamma C_\psi \\ \dot{\psi} = -u_\alpha - S_\psi u_\gamma / e \end{cases} \quad (19)$$

where $e \neq 0$. Moreover, we define $\psi = 0$ and $\dot{\psi} = -u_\alpha$, if $e = 0$.

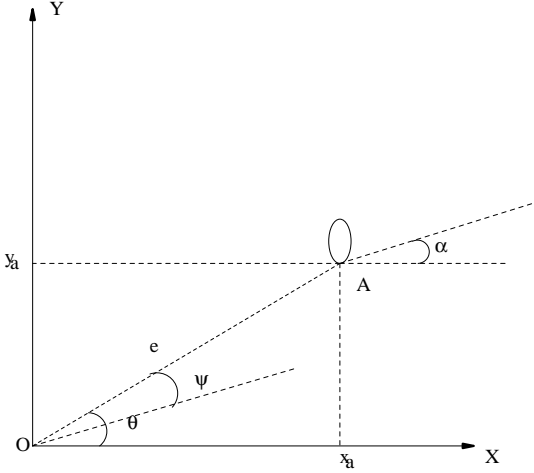

Fig. 4. Parameters in position control

On the basis of the previous considerations, we are now ready to address the aforementioned closed loop steering problem in the following general terms.

Let the robot system be initially located at any non-zero distance from the inertial frame and assume that all state variables required are directly measurable. Then find a suitable, if any, state feedback control law $[u_\alpha, u_\gamma]^T$ which guarantees the state $[e, \beta - \pi/2, \dot{\beta}]$ to be asymptotically driven to the null point $[0,0,0]^T$, while avoiding any attainment of the conditions of $\beta = 0$ (or $\beta = \pi$) in a finite time.

**Proposition 2** Consider the system (18) with the feedback control laws $u_\alpha$ and $u_\gamma$,

$$\begin{cases} u_\alpha = -k_3 Sgn(C_\psi) Sgn(\beta - \pi/2 + \dot{\beta}) \\ u_\gamma = -(k_4 e + u_k) Sgn(C_\psi) \end{cases} \quad (20)$$

where $Sgn$, $k_3$ and $k_4$ are defined in Equations (21) and (22), $k_4$ is a positive scalar constant and $k_4 < k_3 - 1$, which can be designed.

Any state $(e, \beta - \pi/2, \dot{\beta})$ starting from the domain $D$, which is defined as $D = \{(e(0), \beta(0) - \pi/2, \dot{\beta}(0)) \mid e > 0, \ 0 < \beta < \pi,$
$\sqrt{(\beta - \pi/2)^2 + (\beta - \pi/2 + \dot{\beta})^2 / 2} < \pi/2, e, \beta, \dot{\beta} \in R^1\}$ converges to the limiting point $[0,0,0]^T$.

*Proof:*
First, let $Sgn(.)$ be a sign function described as follows:

$$Sgn(x) = \begin{cases} 1, & if \ x \geq 0 \\ -1, & if \ x < 0 \end{cases} \quad (21)$$

Let $k_3 > 2$ be a positive scalar constant, which can be designed and should be less then $\alpha_{max}$. Let

$$\begin{cases} f_1 = |G_m C_\beta + I_m C_\beta S_\beta k_3^2| \\ u_k = (|2(\beta - \pi/2 + \dot{\beta}) + f_1|)/(J_m S_\beta k_3) \end{cases} \quad (22)$$

Set

$$\begin{cases} V = V_1 + V_2 \\ V_1 = \left((\beta - \pi/2)^2 + (\beta - \pi/2 + \dot{\beta})^2\right)/2 \\ V_2 = e^2/2. \end{cases} \quad (23)$$

The time derivative $\dot{V}$ is given by
$$\dot{V} = \dot{V}_1 + \dot{V}_2$$
where
$$\dot{V}_1 = 2(\beta - \pi/2)\dot{\beta} + (\dot{\beta} + \beta - \pi/2)(\dot{\beta} + \ddot{\beta}).$$

Substituting Equations (22) and (20) into Equation (18), we obtain

$$(\dot{\beta} + \beta - \pi/2)\ddot{\beta}$$
$$= -(G_m C_\beta + I_m C_\beta S_\beta)(\dot{\beta} + \beta - \pi/2)$$
$$- |(G_m C_\beta + I_m C_\beta S_\beta)(\dot{\beta} + \beta - \pi/2)|$$
$$- J_m S_\beta k_3 k_4 e|(\dot{\beta} + \beta - \pi/2)| - 2(\dot{\beta} + \beta - \pi/2)^2$$
$$\leq -2(\dot{\beta} + \beta - \pi/2)^2.$$

Such that
$$\dot{V}_1 \leq -(\beta - \pi/2)^2 - \dot{\beta}^2 - (\beta - \pi/2 + \dot{\beta})^2.$$

Thus $V_1$ is positive definite and $\dot{V}_1$ is negative definite. The remaining problem concerns why $\beta$ will not reach 0 or $\pi$ during the entire process.

Since $|\beta - \pi/2| \leq \sqrt{V_1(0)} < \pi/2$ and $V_1$ is monotonically non increasing, they guarantee $0 < \beta < \pi$.

$$\dot{V}_2 = e\dot{e}$$
$$= -eR|C_\psi|(k_4 e + u_k)$$
$$\leq -k_4 e^2 R|C_\psi|$$

Since $R, k_3, k_4, S_\beta$ and $u_k$ are positive, $\dot{V}_2 \leq 0$. In fact, $\dot{V}_2$ is strictly negative, except when

$$\begin{aligned} e &= 0 \\ C_\psi &= 0 \end{aligned} \quad (24)$$

In these cases $\dot{V}_2 = 0$. Equation (24) presents two possible solutions for the system.

Moreover, $C_\psi = 0$ is not a solution, which is evident from Equation (19). By substituting $u_\alpha$ and $u_\gamma$ into Equation (19), we have
$$\dot{\psi} = Sgn(C_\psi)(k_3 Sgn(\beta - \pi/2 + \dot{\beta}) + S_\psi (k_4 e + u_k)/e)$$

Because $\beta \to \pi/2, \dot{\beta} \to 0$ and $C_\beta \to 0, u_k$ will vanish. Thus we obtain
$$\dot{\psi} = Sgn(C_\psi)(k_3 Sgn(\beta - \pi/2 + \dot{\beta}) + k_4 S_\psi)$$

Since $k_3 - 1 > k_4 > 0$, during any sampling period, $\dot{\psi}$ is not zero. Hence, any system trajectory starting from a set in $C_\psi = 0$ will not remain there. Thus, $e = 0$ is the only solution, according to the LaSalle Proposition for nonsmooth systems in (Shevitz, D. & Paden, B. (1994)). Therefore, the proposition is proven.

## 6. Line Tracking Control

For a mobile robot, such as Gyrover, most traveling tasks can be realized by following connected segments of straight lines. For example, Figure 5 shows a mobile robot's travel path along a series of connected corridors.

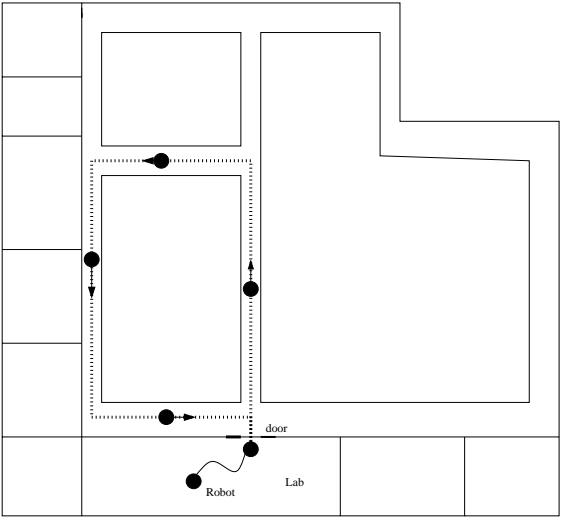

Fig. 5. The robot's path along connected corridors

How to realize the straight line following while keeping the robot vertical is our control problem. As depicted in Figure 6, where $(x_2, y_2)$ is the coordinate of the second point, let us consider Gyrover initially positioned at a neighborhood of the origin (i.e., the start point) and standing almost vertically, about to take a straight line so as to approach the second point.

Let us define $r, e$ and $d$, which are nonnegative, as the distances from A to origin, the line and the second point, respectively. $\theta$ and $\phi$ are defined as in Figure 6. Then, we have the following equations

$$\begin{cases} e\dot{e} = rRu_\gamma S_{\phi-\alpha} S_{\phi-\theta} \\ d\dot{d} = pRu_r \\ p = rC_{\theta-\alpha} - \sqrt{x_2^2 + y_2^2}\, C_{\phi-\alpha} = -dC_{\psi-\alpha} \end{cases}$$

where

$S_{\phi-\alpha} := \sin(\phi-\alpha), S_{\phi-\theta} := \sin(\phi-\theta),$
$C_{\phi-\theta} := \cos(\phi-\theta),$ and $C_{\varphi-\alpha} := \cos(\varphi-\alpha).$

On the basis of previous considerations, we are now ready to address the aforementioned closed loop line tracking problem in general terms.

Let the robot initially locate at some neighborhood the origin and assume that all state variables be directly measurable; then find a suitable (if any) feedback control law $[u_\alpha, u_\gamma]^T$ that guarantees the state $[\beta - \pi/2, \dot{\beta}, e, d]^T$ to be asymptotically driven to the point $[0,0,0,0]^T$, while avoiding any attainment of the conditions of $\beta = 0$ (or $\beta = \pi$) in finite time.

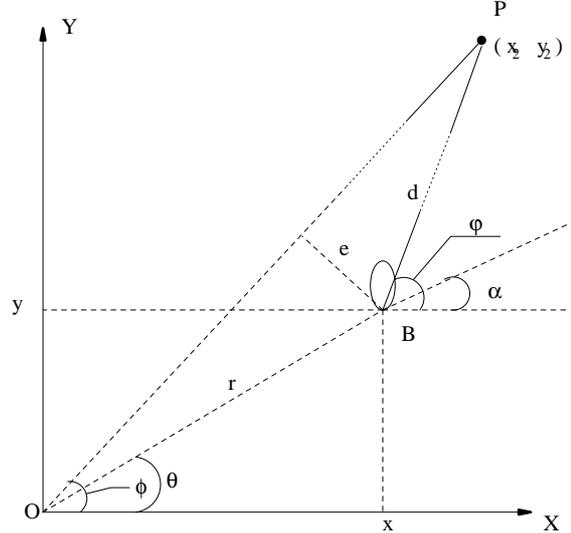

Fig. 6. The parameters in line tracking problem

**Proposition 3** Consider the system (18) with the Feedback control laws $u_\alpha$ and $u_\gamma$,

$$\begin{cases} u_\alpha = -k_3 Sgn(S_{\phi-\alpha} S_{\phi-\theta}) Sgn(\beta - \pi/2 + \dot{\beta}) \\ u_r = -(f_2 + u_k) Sgn(S_{\phi-\alpha} S_{\phi-\theta}) \end{cases} \quad (25)$$

where $f_2$ is defined in Equation (27).
Any state, starting from $[\beta(0) - \pi/2, \dot{\beta}(0), e(0), d(0)]$ with $0 < \beta(0) < \pi$ and $e(0) > 0$, converges to the point $[0,0,0,0]^T$.

*Proof:*
First, we introduce some definitions.
$Sgn(.), k_3$ and $u_k$ are defined as in Equations (21) and (22).
Let $\Theta(.)$ be a sign function described as follows:

$$\Theta(x) = \begin{cases} 1, & \text{if } x \geq o \\ 0, & \text{if } x < 0. \end{cases} \quad (26)$$

We let

$$f_2 = k_5 \Theta(pSgn(S_{\phi-\alpha} S_{\phi-\theta})) \quad (27)$$

where $k_5$ is a positive scalar constant which can be designed and should be less then $\dot{\gamma}_{\max}$.
Set

$$\begin{aligned} V &= V_1 + V_2 + V_3 \\ V_1 &= ((\beta - \pi/2)^2 + (\beta - \pi/2 + \dot{\beta})^2)/2 \\ V_2 &= e^2/2 \\ V_3 &= d^2/2 \end{aligned} \quad (28)$$

The time derivative $\dot{V}$ is given by

$$\dot{V} = \dot{V}_1 + \dot{V}_2 + \dot{V}_3$$

where

$$\dot{V}_1 = (\beta - \pi/2)\dot{\beta} + (\beta - \pi/2 + \dot{\beta})(\dot{\beta} + \ddot{\beta})$$
$$\leq -(\beta - \pi/2)^2 - (\beta - \pi/2 + \dot{\beta})^2$$
$$\dot{V}_2 = e\dot{e} = -rR|S_{\phi-\alpha}S_{\phi-\theta}|(f_2 + u_k)$$
$$\dot{V}_3 = d\dot{d} = pRu_\gamma$$

Since $r, R, f_2$ and $u_k$ are nonnegative, $\dot{V}_2 \leq 0$. This means that the first term $e$ is always non-increasing in time and consequence. If $d = 0$ is maintained, $e$ is also zero. Moreover, from Figure 6, $S_{\phi-\theta} = 0$ and $e = 0$ can be deduced from each other. $\dot{\alpha}$ will never be zero, for $\dot{\alpha} = u_\alpha$. Because $\phi$ is a constant value, it is trivial to know that $e$ will not stop to decrease until $e = 0$.

$V_1$ is positive definite and $\dot{V}_1$ is negative definite. At the beginning $\beta$ is near vertical, so that during the entire process, $\beta \approx \pi/2$ is sustained. That means $\cos\beta \to 0$ and $u_k \approx 0$. By omitting them,

$$\dot{V}_3 = -k_5 pSgn(S_{\phi-\alpha}S_{\phi-\theta})\Theta(pSgn(S_{\phi-\alpha}S_{\phi-\theta})).$$

Thus, $\dot{V}_3$ is negative semi-definite. As with the previous process, it is trivial to prove that the only solution of the system, for $V_3 = 0$ is $d = 0$. Thus, we prove the proposition.

## 7. Experiment

An on-board 100-MHZ 486 computer was installed in Gyrover to deal with on-board sensing and control. A flash PCMCIA card is used as the computer's hard disk and communicates with a stationary PC via a pair of wireless modems. Based on this communication system, we can download the sensor data file from the onboard computer, send supervising commands to Gyrover, and manually control Gyrover through the stationary PC. Moreover, a radio transmitter is installed for a human operator to remotely control Gyrover via the transmitter's two joysticks. One operator uses the transmitter to control the drive speed and tilt angle of Gyrover. Hence, we can record the operator's driving data.

Numerous sensors are installed in Gyrover to measure state variables. Two pulse encoders were
installed to measure the spinning rate of the flywheel and the wheel. Furthermore, we have two gyros and an accelerometer to detect the angular velocity of yaw, pitch, roll, and acceleration respectively. A 2-axis tilt sensor was developed and installed for directly measuring the lean angle and pitch angle of Gyrover. A gyro tilt potentiometer is used to calculate the tilt angle of the flywheel and its rate change.

The on-board computer runs on an OS, called QNX, which is a real-time micro-kernel OS developed by QNX Software System Limited. Gyrover's software system is divided into three main programs: (1) communication server, (2) sensor server, and (3) controller. The communication server is used to communicate between the on-board computer and a stationary personal computer (PC) via an RS232, while the sensor server is used to handle all the sensors and actuators. The controller program implements the control algorithm and communicates among these servers. All programs are run independently in order to allow real-time control of Gyrover. To compensate for the frictions at the joints, we adopt the following approximate mathematical model:

$$F'_f = \mu_v \dot{q} + [\mu_d + (\mu_s - \mu_d)\Gamma]\text{sgn}(\dot{q}) \quad (29)$$

where $\Gamma = diag\left\{e^{(-|\dot{q}_i|/D)}\right\}$, and

$\mu_v \in R^{3\times 3}$ is the viscous friction coefficient;

$\mu_d \in R^{3\times 3}$ is the dynamic friction coefficient;

$\mu_s \in R^{3\times 3}$ is the static friction coefficient.

We have $\mu_v = diag\{0.17, 0.15, 0.09\}$,
$\mu_d = diag\{0.1, 0.1, 0.07\}, \mu_s = diag\{0.3, 0.25, 0.1\}$.

### 7.1. Balance Control

The purpose of this set of experiments is to keep Gyrover balance. Some experimental results are shown in Figure 7 and Figure 8.

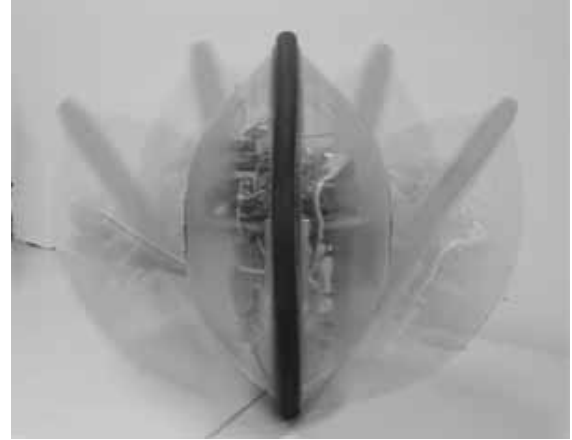

Fig. 7. Camera picture in balance control

Despite several successful examples, control laws sometimes failed. We believe this is because the initial condition already went out to the domain $D$ in Proposition 1. With regard to balance control, the initial condition seems to be too narrow for the constraint, $\sigma < \pi/2$. If we make a small modification to $u_6$ as follows, the controller can be used in a wider range of initial conditions.

$$u_6 = -((2+k_1)(\beta-\pi/2) + (3+2k_1)\dot{\beta} + (2+k_1)\ddot{\beta}$$
$$+ h_1(t)\dot{\beta} + h_2(t)u_5 )/h_3(t)$$

where $k_1$ is a positive number, which can be designed.

To prove the subsystem $\beta, \dot{\beta}, \ddot{\beta}$ is asymptotically stabilized, we consider the following Lyapunov function candidate

$$V^* = (\beta - \pi/2)^2/2 + (\dot{\beta} + \beta - \pi/2)^2/2$$
$$+ (\ddot{\beta} + (1+k_1)(\dot{\beta} + \beta - \pi/2))^2/2$$

Its derivative is

$$\dot{V}^* = -(\beta - \pi/2)^2 - k_1(\dot{\beta} + \beta - \pi/2)^2 \\ - (\ddot{\beta} + (1+k_1)(\dot{\beta} + \beta - \pi/2))^2.$$

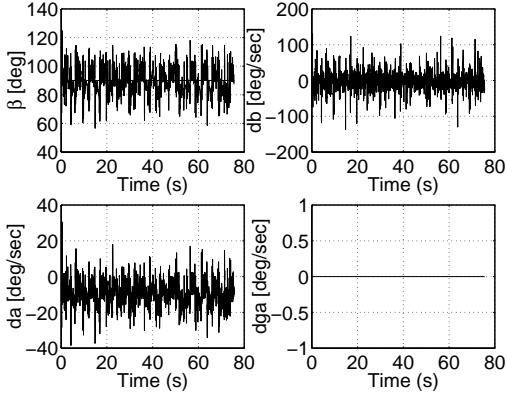

Fig. 8 Sensor data in the balance control

### 7.2. Position Control

In this experiment, Gyrover is required to move from a Cartesian space point $(x_0, y_0)$ to the original point of the Cartesian space, where $x_0 = 3$ m and $y_0 = 4$ m. We mount a high resolution $2/3''$ format CANON COMMUNICATION CAMERA VC-C1 on a tripod. The camera had been calibrated and we have mapped the field of vision to the Cartesian space coordinate which is anchored on the ground. The camera has a pixel array with $768(H) \times 576(V)$. In order to communicate the data, an interface board (digital I/O) is installed in a PC and there are wireless modems to connect the PC and Gyrover. The first problem experienced was that the system states exhibited highly oscillatory behavior and the second was that control inputs sometimes needed to switch too sharply and fast. Both will cause difficulties for real time control and result in worse performance. To solve these problems, we propose to replace the sign functions in the controllers with tanh functions.

Let $Tanh(.)$ be a bipolar function described as follows:

$$Tanh(x) = \frac{1 - e^{-k_6 x}}{1 + e^{-k_6 x}} \quad (30)$$

where $k_6$ is a positive scalar constant, which can be designed.

Let $Uanh(.)$ be a unipolar function described as follows:

$$Uanh(x) = \frac{1}{1 + e^{-k_7 x}} \quad (31)$$

where $k_7$ is a positive scalar constant, which can be designed.

We substitute $Sgn(.)$ and $\Theta(.)$ with $Tanh(.)$ and, respectively. However, that they have different values is a problem, if $x = 0$. In a real time experiment, $x = 0$ very seldom appears and can not be maintained, because there is too much noise. Thus, it is safe to perform the suggested substitution from this point of view. The trajectory that Gyrover had traveled is shown in Figure 9. A number of sensor reading results are shown in Figure 10.

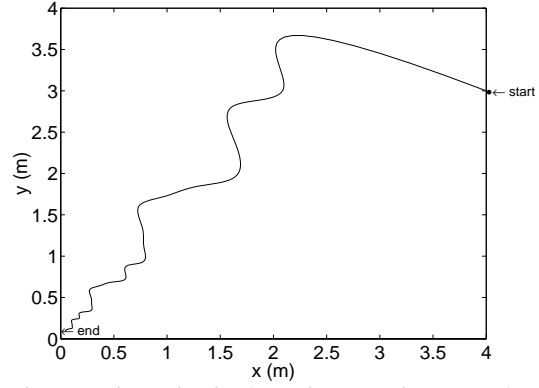

Fig. 9 Trajectories in the point-to-point control

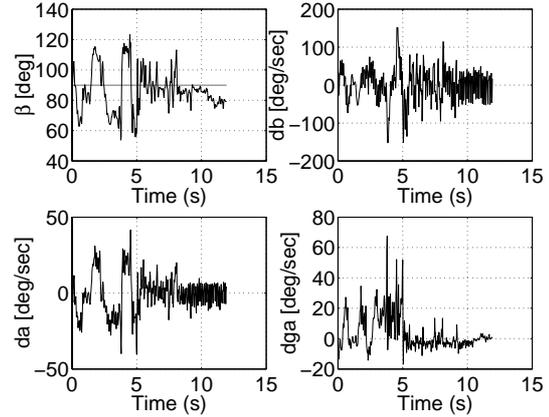

Fig. 10. Sensor data in the point-to-point control

### 7.3. Line Tracking

In this experiment, Gyrover is required to travel a straight path which is about 5 m long. The trajectory that Gyrover traveled is shown in Figure 12. A number of sensor reading results are shown in Figure 13.

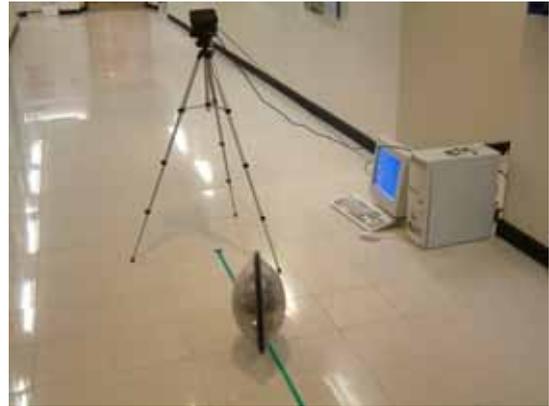

Fig. 11. Experiment in line following control

## 8. Conclusion

In this paper, we studied control problems for a single wheel, gyroscopically stabilized robot. We investigated the dynamics of the robot system, and analyzed the two classes of nonholonomic constraints of the robot. We proposed three control laws for balance, point-to-point control and line tracking. The three problems considered are fundamental tasks for Gyrover control.

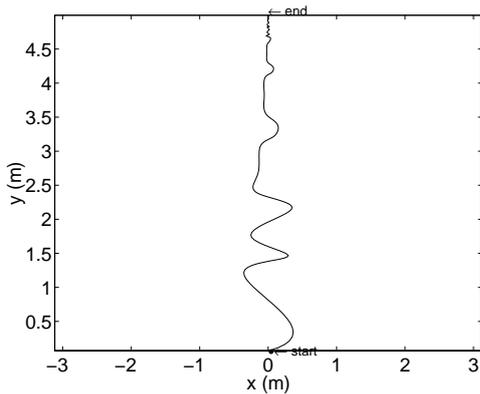

Fig. 12. Trajectories in the straight path test

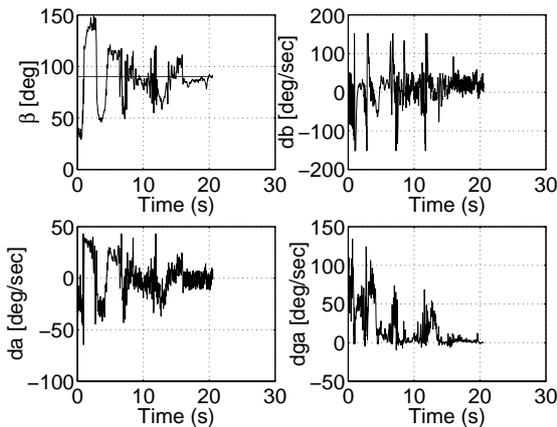

Fig.12. Sensor data in the straight path test

**Acknowledgment**

This work is supported in part by Hong Kong Research Grant Council under the grants CUHK 4403/99E, CUHK 4228/01E and Hong Kong SAR Government under grand ITS/140/01.